\documentclass[10pt,twocolumn,letterpaper]{article}

\usepackage{cvpr}              

\usepackage{array}    
\usepackage{booktabs} 
\usepackage{graphicx} 
\usepackage{algorithm}
\usepackage{algpseudocode}
\usepackage{listings}
\usepackage{xcolor}
\usepackage{makecell}
\usepackage[utf8]{inputenc}
\usepackage{tikz}
\usetikzlibrary{shapes, shadows, positioning, backgrounds}
\usepackage{tabularx}

\definecolor{cvprblue}{rgb}{0.21,0.49,0.74}
\usepackage[pagebackref,breaklinks,colorlinks,allcolors=cvprblue]{hyperref}

\def\paperID{CVPR-NAS26-15} 

\def\confName{CVPR}
\def\confYear{2026}

\title{LEMUR 2: Unlocking Neural Network Diversity for AI}

\author{Tolgay Atinc Uzun$^{*}$, \hspace{0.05cm} Waleed Khalid, \hspace{0.05cm} Saif U Din, \hspace{0.05cm} Sai Revanth Mulukuledu, \hspace{0.05cm}
Akashdeep Singh,\\ Chandini Vysyaraju, \hspace{0.05cm} Raghuvir Duvvuri, \hspace{0.05cm} Avi Goyal, \hspace{0.05cm}
Yashkumar Rajeshbhai Lukhi,\\Muhammad A. Hussain, \hspace{0.00cm} Krunal Jesani,\hspace{0.00cm} Usha Shrestha,\hspace{0.00cm}
Yash Mittal,\hspace{0.00cm} Roman Kochnev,\hspace{0.00cm} Pritam Kadam,\\ Mohsin Ikram, \hspace{0.05cm}  
Harsh R. Moradiya, \hspace{0.05cm} Alice Arslanian, \hspace{0.05cm} Dmitry Ignatov$^{\dagger}$, \hspace{0.05cm} Radu Timofte\\
\small{Computer Vision Chair, University of W\"urzburg, Germany}\\
{\tt\small $^*$t.atincuzun@gmail.com, $^{\dagger}$dmytro.ignatov@uni-wuerzburg.de}\\
\href{https://github.com/ABrain-One/NN-Dataset}{https://github.com/ABrain-One/NN-Dataset}}

\begin{document}
\maketitle
\begin{abstract}
Existing NAS benchmarks (e.g., NAS-Bench, NATS-Bench) cover only narrow, task-specific regions of the architectural design space and lack cross-domain or deployment-aware evaluation. LEMUR 2 introduces a large-scale, extensible framework unifying generative, evaluative, and deployment pipelines to unlock neural-network diversity. It comprises over 14{,}000 distinct architectures and more than 750{,}000 structured training records documenting model performance, hyperparameters, and task outcomes. These models were produced through AST-based code mutation, genetic and reinforcement-learning evolution, generation of fractal architectures, and synthesis guided by a Large Language Model (LLM). This includes deep models generated with the retrieval-augmented system NN-RAG, which derived and used architectural motifs from over 900 PyTorch modules extracted from public repositories. LEMUR 2 further employs $\text{NN-VR}$ and $\text{NN-Lite}$ pipelines for automated deployment and latency benchmarking on heterogeneous mobile and Unity-based VR platforms, providing real-device performance metadata. It spans multimodal tasks—image captioning, text-to-image synthesis, and language modeling—supporting cross-domain analysis of architectural transferability. By linking diverse architectures, tasks, and deployment data, LEMUR 2 provides the data foundation for LLM fine-tuning and coupling diverse architectural origins with large-scale, cross-platform empirical validation. This dataset defines a new basis for reproducible and data-driven AI design, advancing the emerging paradigm of LLM-driven AutoML and architectural generalization across modalities and hardware.
\vspace{-0.2cm}

\end{abstract}    
\section{Introduction}
\label{sec:intro}

Neural networks underpin numerous breakthroughs in artificial intelligence, delivering state-of-the-art results in fields such as computer vision and natural language processing. Their growing complexity, however, has exposed the limits of manual design. In response, the field is shifting toward a paradigm centered on creating and leveraging large-scale, reusable collections of neural network models. The goal of this shift is to replace manual intuition with data-driven discovery, enabling researchers to identify latent patterns, structural regularities, and generalizable design principles from diverse model corpora. Such model-centric datasets are now critical resources for benchmarking, neural architecture analysis, meta-learning, and automated machine learning.

Most publicly available neural architecture repositories are built around a single construction paradigm, typically exhaustive or near-exhaustive enumeration of a fixed cell-based search space. This yields methodologically homogeneous collections that cover only a narrow region of the architectural design space and are not readily extensible to models obtained by other means. At the same time, these repositories rarely provide an end-to-end path from architectural specification to task-level evaluation and device-level deployment, making it difficult to study accuracy-latency trade-offs or reuse models across modalities.

We propose multiple strategies that build on the LEMUR~\cite{goodarzi2025lemurneuralnetworkdataset} dataset and accompanying pipeline to improve architectural diversity and operational integration. To generate more models, we employ programmatic editing, reinforcement learning, evolution-based search, fractal construction, retrieval-augmented block extraction, and LLM-driven synthesis. On the deployment side, we provide NN-Lite and NN-VR for benchmarking, testing, and verification across multiple tasks on resource-constrained targets under a shared protocol. Statistics are available in the project repository\footnote{Github Repository: \scriptsize \url{https://github.com/ABrain-One/nn-dataset}}.


\subsection{Related Work}

In neural network research, datasets are the backbone of training, evaluation, and benchmarking. There are plentiful datasets in image, text, and domain-specific corpora \cite{lin2014microsoft} that provide large, labeled collections of data. By contrast, resources that systematically capture the structures, configurations, and resulting performance of neural networks themselves are far less common.

\textbf{Dataset of models} The closest line of work comes from the neural architecture search (NAS) community. Benchmarks such as NAS-Bench-101~\cite{ying2019bench} and its extension NAS-Bench-201~\cite{dong2020nasbench201} enumerate a fixed, small search space of convolutional cells and store their training and evaluation results to enable reproducible comparison of NAS algorithms. NATS-Bench~\cite{dong2021nats} generalizes this idea to both topology and size spaces across multiple datasets, while TransNAS-Bench-101~\cite{duan2021transnas} expands it to several vision-style tasks to study transferability. NAS-Bench-NLP~\cite{klyuchnikov2022bench} moves beyond vision to language models, HW-NAS-Bench~\cite{li2021hw} adds hardware and latency measurements, and JAHS-Bench-201~\cite{Bansal2022JAHSBench201} couples architectures with hyperparameters for joint optimization. 

\textbf{AutoML Frameworks and Model Repositories} AutoML frameworks such as AutoKeras~\cite{jin2019auto} and TPOT~\cite{pmlr-v64-olson_tpot_2016} provide tooling for automated model selection and hyperparameter optimization but do not maintain large-scale, curated repositories of architectures with performance metadata. Similarly, model zoos like TensorFlow Hub~\cite{tensorflowhub} and PyTorch Hub~\cite{pytorchhub} offer pre-trained models but lack standardized evaluation across tasks and hardware, and they do not support systematic architecture generation or diversity analysis.

\subsection{Our Contribution}

This paper presents LEMUR 2, an extension of the original LEMUR~\cite{goodarzi2025lemurneuralnetworkdataset} framework. While LEMUR served as a repository for existing neural networks and their associated statistics, inspired by recent advancements in the use of LLMs across various domains~\cite{kochnev2025optunavscodellama,Gado2025llm,Rupani2025llm}, LEMUR 2 is designed to include generative systems that leverage existing data for the automated creation of new network architectures~\cite{ABrain.NNGPT} as well as an assessment tool for testing on edge devices~\cite{ABrain.NN-Lite} and deployment on different platforms. 

First, the dataset was expanded to over 14,000 models using several distinct, automated generation methodologies. These methods include genetic algorithms, reinforcement learning-based layer masking for network mutation, abstract syntax tree (AST) editing of model source code, and fractal-inspired generation. To facilitate dynamic collection, we also introduce NN-RAG, a retrieval-augmented system that extracts validated, self-contained PyTorch~\cite{PyTorch} modules from external codebases, creating a library of over 900 reusable components. The dataset is further enriched with metadata from a systematic evaluation of 6,000 data transformation pipelines, providing guidance on optimal data preprocessing. The framework's applicability is demonstrated through the inclusion of additional tasks, such as image captioning, text-to-text and text-to-image synthesis.

Second, to connect theoretical metrics with practical performance, we added deployment-aware metadata. We built NN-Lite, an automated pipeline that converts, deploys, and benchmarks PyTorch models on the Android platform. This system processed over 7,500 models, populating the LEMUR 2 database with their on-device inference latencies, empirical measurements often absent from standard benchmarks. This deployment analysis was extended to immersive applications with NN-VR, a system for the automated conversion and performance evaluation of models within the Unity engine for virtual reality contexts.

Collectively, these contributions establish LEMUR 2 as a public research resource. It provides a large-scale architectural corpus, a set of generative methodologies, and multi-platform deployment benchmarks that support research in automated AI design. The empirical value of the LEMUR dataset, as demonstrated within the NNGPT project, is proved by its successful application to LLM-guided generative synthesis, exploration of complex architectural structures, and improved generalization across multiple computer vision tasks~\cite{kochnev2025optunavscodellama, ABrain.Architect, mittal2025fractal, ABrain.Transform, ABrain.NN-Captioning_2025, ABrain.Prompt,ABrain.CV_Channel,ABrain.Feedback_Memory}.

\begin{table*}[ht]
\centering
\caption{Comparison of prominent NAS benchmarks. \#Entries denotes the number of distinct queryable model specifications released by the benchmark (architectures, or architecture+HP configurations), each with stored trained or predicted metrics. Multi-Task indicates whether multiple datasets and/or multiple tasks are included within the benchmark's domain.}
\label{tab:nas_comparison_comprehensive}
\resizebox{\textwidth}{!}{%
\begin{tabular}{@{}l l l r l l l c@{}}
\toprule
& & & & & & \multicolumn{1}{c}{} & \multicolumn{1}{c}{\textbf{Capabilities}} \\
\cmidrule(l){8-8}
\textbf{Benchmark} & \textbf{Domain} & \textbf{Search Space} & \textbf{\#Entries} & \textbf{Datasets} & \textbf{Tasks} & \textbf{Evaluation} & \textbf{Multi-Task} \\
\midrule
\textbf{NAS-Bench-101} \cite{ying2019bench} & Vision & Cell tabular & 423,624 & CIFAR-10 & Image classification & Trained & $\times$ \\
\textbf{NAS-Bench-201} \cite{dong2020nasbench201} & Vision & Cell tabular & 15,625 & CIFAR-10, CIFAR-100, ImageNet16-120 & Image classification & Trained & \checkmark \\
\textbf{NAS-Bench-301} \cite{siems2020nasbench301} & Vision & Cell surrogate (DARTS) & $\sim$60,000 & CIFAR-10 & Image classification & Predictive & $\times$ \\
\textbf{NATS-Bench} \cite{dong2021nats} & Vision & Topology+size tabular & 48,393 & CIFAR-10, CIFAR-100, ImageNet16-120 & Image classification & Trained & \checkmark \\
\textbf{TransNAS-Bench} \cite{duan2021transnas} & Vision & Cell+macro tabular & 7,352 & Taskonomy (subset) & 7 vision tasks & Trained & \checkmark \\
\textbf{NAS-Bench-NLP} \cite{klyuchnikov2022bench} & NLP & RNN tabular & 14,322 & PTB; WikiText-2 (subset) & Language modeling & Trained & \checkmark \\
\textbf{NAS-Bench-Graph} \cite{qin2022bench} & Graph & GNN tabular & 26,206 &
\begin{tabular}[t]{@{}l@{}}
Cora, CiteSeer, PubMed \\
Coauthor-CS/Physics, Amazon-Photo/Computers \\
ogbn-arXiv, ogbn-proteins
\end{tabular}
& Node classification & Trained & \checkmark \\
\textbf{NAS-Bench-ASR} \cite{mehrotra2021nasbenchasr} & Audio & ASR cell tabular & 8,242 & TIMIT & Speech recognition & Trained & $\times$ \\
\textbf{HW-NAS-Bench} \cite{li2021hw} & Hardware & NB201 + FBNet (HW) & 15,625 & CIFAR-10, CIFAR-100, ImageNet16-120 & HW cost (latency+energy; 6 devices) & Trained + Measured & \checkmark \\
\textbf{NAS-Bench-1Shot1} \cite{zela2020nasbench1shot1} & Method & One-shot tabular subsets & 399,048 & CIFAR-10 & Image classification & Trained & $\times$ \\
\textbf{JAHS-Bench-201} \cite{Bansal2022JAHSBench201} & HPO & Arch+HP surrogate & 270,000 & CIFAR-10, Fashion-MNIST, Colorectal Histology & Image classification + HPO & Predictive & \checkmark \\
\midrule
\textbf{LEMUR 2 (Ours)} & \textbf{Universal} & \textbf{Open code (LLM)} & \textbf{$>$14,000} & \textbf{Multiple (user-defined)} & \textbf{Multi-domain} & \textbf{Trained} & \checkmark \\
\bottomrule
\end{tabular}%
}
\end{table*}

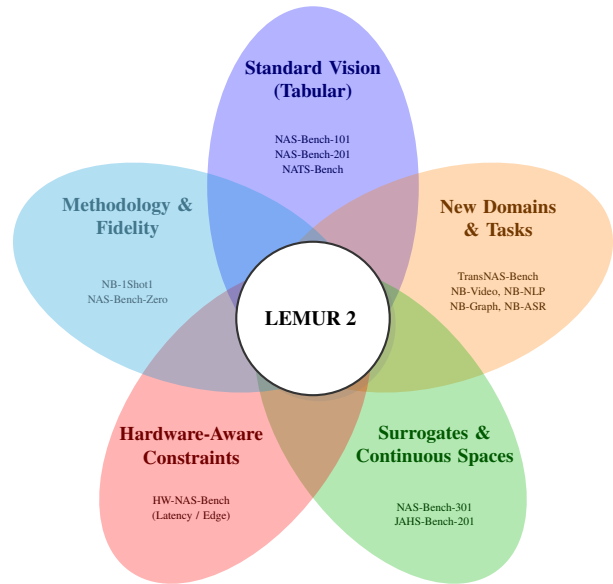
\begin{figure}[ht]
\centering
\begin{tikzpicture}[scale=0.70, transform shape]

    \tikzset{
        petal/.style={
            ellipse,
            minimum width=4.2cm, 
            minimum height=7.5cm,
            draw=none,
            fill opacity=0.3,
            text opacity=1,
            align=center,
            font=\sffamily
        },
        header/.style={
            font=\bfseries\fontsize{10.4pt}{12.5pt}\selectfont,
            text width=3.6cm,
            align=center,
            anchor=center
        },
        list/.style={
            font=\fontsize{6.1pt}{8pt}\selectfont,
            text width=3.4cm,
            align=center,
            anchor=center
        }
    }

    \draw[fill=blue, petal] (0,2.5) ellipse (2.0cm and 3.4cm);
    \node[header, text=blue!45!black] at (0, 4.5) {Standard Vision\\(Tabular)};
    \node[list, text=blue!35!black] at (0, 3.1) {
        NAS-Bench-101\\
        NAS-Bench-201\\
        NATS-Bench
    };

    \draw[fill=orange, petal, rotate=-72] (0,2.5) ellipse (2.0cm and 3.4cm);
    \node[header, text=orange!45!black] at (3.5, 1.9) {New Domains\\\& Tasks};
    \node[list, text=orange!35!black] at (3.5, 0.5) {
        TransNAS-Bench\\
        NB-Video, NB-NLP\\
        NB-Graph, NB-ASR
    };

    \draw[fill=green!70!black, petal, rotate=-144] (0,2.5) ellipse (2.0cm and 3.4cm);
    \node[header, text=green!35!black] at (2.3, -2.4) {Surrogates \&\\Continuous Spaces};
    \node[list, text=green!25!black] at (2.3, -3.7) {
        NAS-Bench-301\\
        JAHS-Bench-201
    };

    \draw[fill=red, petal, rotate=-216] (0,2.5) ellipse (2.0cm and 3.4cm);
    \node[header, text=red!45!black] at (-2.3, -2.4) {Hardware-Aware\\Constraints};
    \node[list, text=red!35!black] at (-2.3, -3.6) {
        HW-NAS-Bench\\
        (Latency / Edge)
    };

    \draw[fill=cyan!80!blue, petal, rotate=-288] (0,2.5) ellipse (2.0cm and 3.4cm);
    \node[header, text=cyan!45!black] at (-3.5, 1.9) {Methodology \&\\Fidelity};
    \node[list, text=cyan!35!black] at (-3.5, 0.5) {
        NB-1Shot1\\
        NAS-Bench-Zero
    };

    \node[circle, 
          fill=white, 
          draw=black!80, 
          thick,
          minimum size=2.9cm,
          align=center,
          drop shadow={opacity=0.2}] at (0,0) {
            \textbf{\fontsize{11pt}{13pt}\selectfont LEMUR 2}
          };

\end{tikzpicture}
\caption{
Taxonomy of NAS benchmarks organized by their primary differentiating dimension.
\textbf{LEMUR 2} lies at the intersection, providing a unified generative framework that spans tabular evaluation, methodological abstractions, surrogate modeling, hardware awareness, and multi-domain tasks.
}
\vspace{-4.2mm}

\end{figure}

\section{Methodology}

We extend the LEMUR~\cite{goodarzi2025lemurneuralnetworkdataset} corpus as a modular generation and integration framework built on the original experiment manager for task execution and result persistence. Corpus access and downstream processing use the \texttt{data} interface, which exposes stored records as a fixed-schema pandas \texttt{DataFrame} for uniform filtering and aggregation across tasks, datasets, metrics, architectures, epochs, accuracy, runtime, parameter counts, and transformation identifiers without direct database queries. On this layer, self-contained extension modules generate candidate model definitions or transformation programs with metadata and submit them to the standard LEMUR pipeline for evaluation. These modules combine LLM-based generation from natural-language guidance with programmatic generators that produce valid variants faster and at lower cost, enabling scalable, heterogeneous expansion under a shared reporting format.

Architectural diversity in LEMUR 2 comes from combining heterogeneous generators and sources, so the corpus is not centered on a single backbone family. For generators that operate by mutating a seed architecture, we instantiate them on a lightweight reference template supported by the generator to enable high-throughput evaluation and controlled attribution of operator effects, while still producing many structurally distinct variants.




\subsection{Task Extensions}

\subsubsection{Image Captioning}

NN-Caption is an automated image-captioning pipeline built on LEMUR in which an LLM, iteratively prompted with a baseline ResNet--LSTM captioner and code templates for the LEMUR \texttt{Net} interface (\texttt{\_\_init\_\_}, \texttt{train\_setup}, \texttt{learn}, \texttt{forward}), generates PyTorch encoder--decoder architectures that couple a convolutional image encoder with sequence decoders (LSTM, GRU~\cite{chung2014empirical}, Transformer). Generated code is normalized, validated via Python's abstract syntax tree, integrated into LEMUR, compiled, and trained on MS~COCO~\cite{lin2014microsoft}, with caption quality evaluated using BLEU-4~\cite{10.3115/1073083.1073135}; in total, NN-Caption contributes 357 captioning architectures to LEMUR~2.


\subsubsection{Text-to-Image}

LEMUR includes a text-to-image pipeline spanning three generative families: a diffusion model (UNet-D)~\cite{rombach2022high,ho2020denoising,ronneberger2015u}, a GAN~\cite{zhang2019self}, and a CVAE-GAN~\cite{kingma2013auto}, all conditioned on natural-language text within a unified training and evaluation framework. For diffusion, a pre-trained CLIP text encoder~\cite{radford2021learning} provides embeddings that are fused with time-step embeddings and injected into a multi-scale U-Net encoder--decoder with residual and attention blocks, optimized with a denoising diffusion objective. The GAN uses a generator that fuses text representations with a convolutional upsampling stack and a discriminator operating on joint image--text pairs, whereas the CVAE-GAN employs a conditional VAE whose decoder serves as the generator alongside an adversarial discriminator that promotes sharp outputs. All models are trained on paired text--image data with architecture-specific losses (diffusion, adversarial, or variational plus adversarial), and performance is quantified via CLIP-based text--image similarity between generated images and their prompts.

\subsubsection{Text-to-Text}

LEMUR is extended to natural language generation via a text-to-text pipeline for training and evaluating recurrent language models with standardized metrics. A lightweight WikiText data loader tokenizes text into integer sequences and forms batched streams for truncated backpropagation through time~\cite{58337}, with context length, vocabulary size, and batching strategy configured via a shared dictionary. Two recurrent baselines are instantiated: a stacked Elman RNN~\cite{elman1990finding} with learned token embeddings, \texttt{tanh} activations, and a linear next-token decoder, and a stacked LSTM~\cite{hochreiter1997long} with input, forget, and output gates that improve stability and long-range dependency modeling. Evaluation uses perplexity and BLEU, with all LEMUR metrics normalized to $[0,1]$; BLEU is computed at corpus level with smoothed $n$-gram precision on the validation set.


\subsubsection{Mixture-of-Experts (MoE)}

Mixture-of-Experts architectures are integrated into LEMUR to study expert composition and routing strategies on CIFAR-10~\cite{krizhevsky2009learning}. Eight variants are constructed in three stages: (i) homogeneous MoEs based on a single backbone type, (ii) AlexNet-based MoEs, and (iii) a heterogeneous MoE combining multiple backbones.

Homogeneous configurations use identical expert architectures and a sparse routing mechanism. For each input, a gating network selects the two experts with the highest scores (top-2 routing), and only these experts are evaluated. Heterogeneous configurations combine four different backbone architectures—AlexNet~\cite{NIPS2012_c399862d}, AirNet~\cite{chee2018airnet}, DenseNet~\cite{huang2017densely}, and BagNet~\cite{brendel2019approximating}—and employ a soft routing strategy in which the gating network produces a weight for each expert. The final prediction is obtained as a weighted aggregation of the experts’ outputs. All MoE variants are trained and evaluated within the LEMUR framework using the standard CIFAR-10 training protocol.

\subsection{Achieving Diversity}

\subsubsection{AST-Based Mutations}

An AST-based mutation mechanism\cite{ABrain.CV_Channel} generates structurally consistent architectural variants by modifying the source code of existing models while operating on the channel dimensions of convolutional and linear layers and preserving tensor-shape compatibility. A symbolic tracer built on \texttt{torch.fx}~\cite{reed2022torch} constructs a computation graph whose nodes represent layers with explicit dataflow edges, and maintains a source map from each module instantiation to its file location. Given a target layer, a planning component uses this graph to identify all dependent consumer layers and infer required input/output channel adjustments, including in architectures with residual connections. The mutation plan is applied by parsing the implementation into an AST, locating the corresponding constructor calls via the recorded coordinates, and updating channel-related arguments; the modified AST is then rendered back to source code, re-imported, and each mutant is validated via instantiation and forward/backward passes on dummy inputs. This procedure yields approximately 1000 additional architectures in the corpus.

\subsubsection{Reinforcement Learning}

The LEMUR dataset is processed by programmatically parsing each network's source code and identifying layer-instantiation statements (e.g., \texttt{nn.Conv2d}, \texttt{nn.Linear}) within the \texttt{\_\_init\_\_} method. These code blocks are removed and replaced with placeholder tokens, while preserving the surrounding class structure, method signatures, and forward pass logic. This produces a dataset of masked model skeletons paired with their complete source code as ground truth for training.

The reinforcement learning pipeline then iteratively refines architecture generation by completing these masked skeletons. The LLM acts as a policy, generating code completions (actions) given the masked skeleton (state). Each generated model undergoes validation checks for compilation and tensor shape consistency via a dummy forward pass. Successful models are trained for one mini-epoch on CIFAR-10 to measure accuracy.

A composite reward is calculated: $-1.0$ for validation failure, $+0.2$ for each passed stage (compilation, forward pass, training), and $+1.0 \times \Delta accuracy$  for improvement over baseline. This reward updates the LLM's policy using Group Relative Policy Optimization (GRPO)~\cite{shao2024deepseekmath}. 

\begin{figure}
    \centering
    \includegraphics[width=1\linewidth]{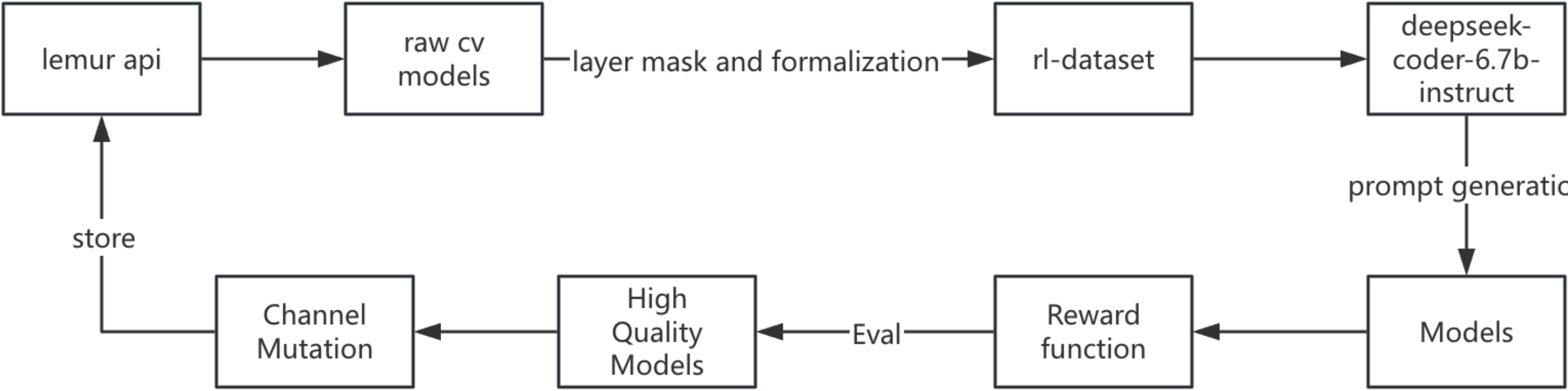}
    \caption{Reinforcement Learning pipeline overview. Using the data from LEMUR, the networks are masked to construct a training dataset. By deriving the policy from LLM, generated sequences of models are assessed and rewarded.}
    \label{fig:rl_pipeline}
    \vspace{-3.2mm}

\end{figure}

\subsubsection{Genetic Algorithm}

A genetic algorithm explores AlexNet-style architectures on CIFAR-10~\cite{krizhevsky2009learning} using a parameterized representation that includes architectural and training hyperparameters. Populations of candidate networks are evolved over generations via crossover and mutation on these parameter vectors; fitness is approximated by training each model for a small number of epochs and using validation accuracy for selection, with a checksum over the architecture description used to discard duplicates and maintain diversity. Two search spaces are considered: one restricted to hyperparameters (e.g., filter counts, kernel sizes, learning rate, dropout), and one that additionally includes block-level structural choices (pooling type, activation function, batch normalization). Across both spaces, the algorithm yields approximately 2{,}000 distinct AlexNet-type architectures.

\begin{figure}
    \centering
    \includegraphics[width=0.5\linewidth]{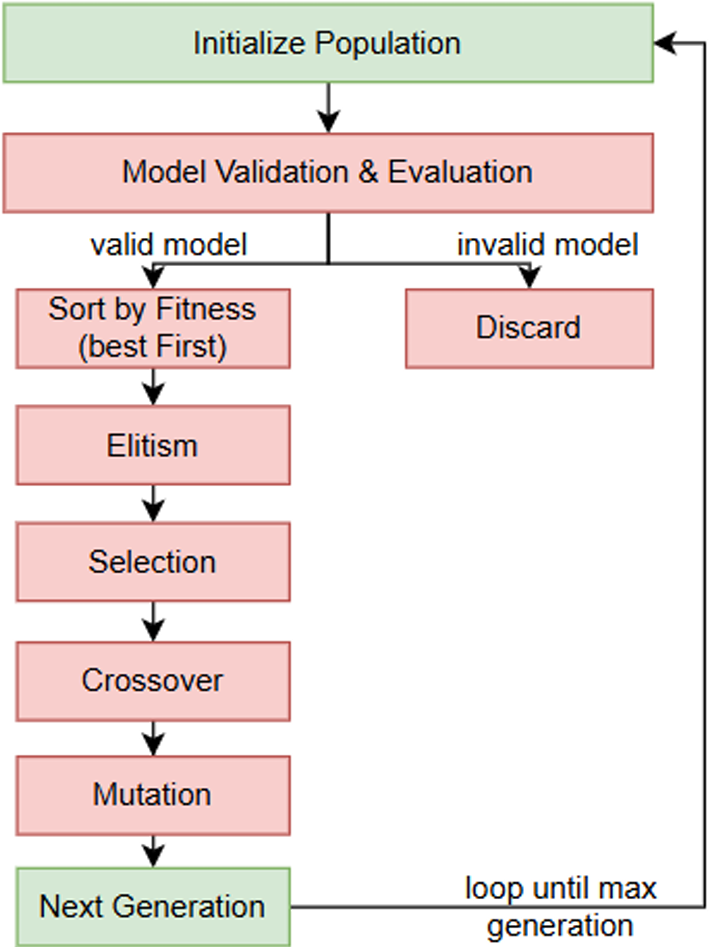}
    \caption{Genetic algorithm model pipeline overview.}
    \label{fig:genetic_pipeline}
    \vspace{-3.2mm}

\end{figure}

\subsubsection{Fractal-Inspired Computational Architectures}

LEMUR includes a FractalNet-style~\cite{larsson2016fractalnet} generator~\cite{mittal2025fractal} that constructs self-similar, recursively defined multi-column networks whose topology is governed by fractal depth $N$ (recursion level) and column width (number of parallel pathways), enabling balanced growth in depth and width. The pipeline comprises (i) configuration generation, which systematically samples architectural blueprints via permutations of convolution, normalization, activation, and dropout layers; (ii) template-based model instantiation, which programmatically realizes the specified recursive, multi-column PyTorch architectures; and (iii) automated evaluation, which executes a standardized training and logging protocol. To scale to over 1{,}200 unique architectures, training employs Automatic Mixed Precision (AMP) to accelerate computation with half-precision arithmetic and gradient checkpointing to reduce GPU memory via recomputation of intermediate activations.

\subsubsection{Retrieval-Augmented Generation}

The NN-RAG component~\cite{ABrain.NN-RAG, ABrain.Architect} constructs a library of reusable PyTorch modules by mining existing codebases. It scans repositories for subclasses of \texttt{torch.nn.Module} that implement a \texttt{forward()} method and extracts them as self-contained units that preserve the original imports and semantics.

Source files are parsed using LibCST~\cite{libcst}, which maintains syntactic structure, formatting, and comments. A scope-sensitive dependency resolver computes, for each candidate module, the minimal transitive closure of required definitions (such as auxiliary classes, functions, and constants) without executing the code. The resulting set of definitions is topologically ordered to ensure definition-before-use and is written out as an isolated module. Each extracted module passes through a multi-stage validation pipeline comprising abstract syntax tree parsing, bytecode compilation, and sandboxed execution to detect syntax and import-time errors. Valid modules are then integrated into LEMUR as reusable building blocks, with associated metadata linking them to their source repositories.


\begin{figure}
    \centering
    \includegraphics[width=1\linewidth]{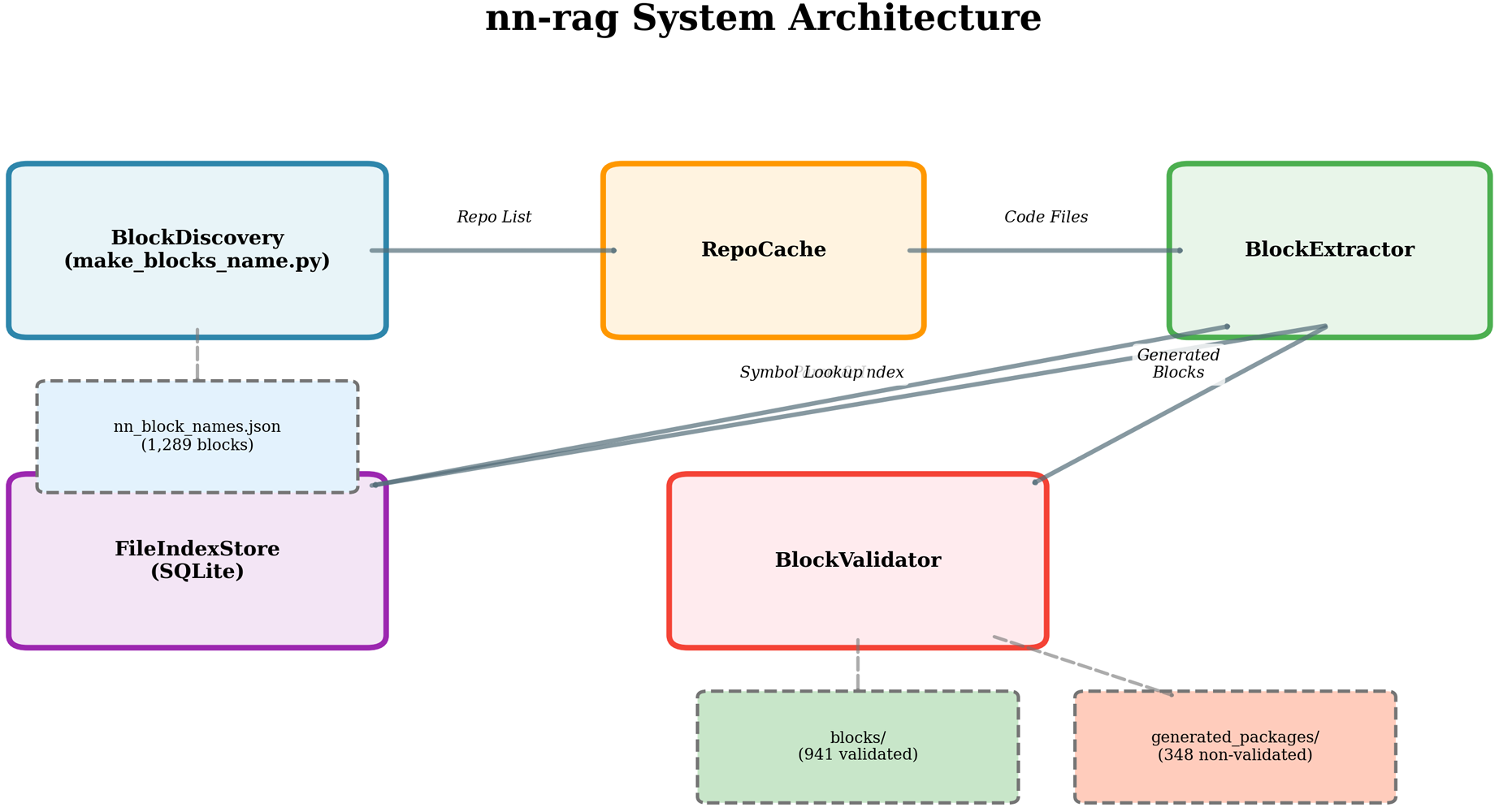}
    \caption{NN-RAG pipeline overview.}
    \label{fig:nnrag_pipeline_simple}
    \vspace{-3.2mm}

\end{figure}

\subsubsection{Few-Shot Architecture Prompting}

To improve the stability of LLM-based architecture generation, we employ few-shot prompting and deduplication. Prompts include a small set of high-performing LEMUR architectures as exemplars, together with a description of the target dataset, and the LLM is instructed to produce a new architecture consistent with the LEMUR interface; the number of exemplars is treated as a tunable parameter. Each generated architecture is normalized into a canonical form and assigned a whitespace-stripped MD5 hash, which serves as a database key for fast detection and removal of exact duplicates prior to training. Non-duplicate architectures are trained under the standard LEMUR protocol and their performance recorded. For statistical analysis, evaluation metrics are aggregated in a dataset-balanced manner by first computing performance per dataset and then averaging across datasets to avoid bias from uneven task coverage.

\subsubsection{Data Transformations}

In line with prior work~\cite{Aboudeshish2025augmentation}, and with the objective of diversifying data augmentation within the corpus using the implementation described in~\cite{ABrain.Transform}, transformation pipelines are generated and evaluated on the CIFAR-10 dataset. A fixed ResNet architecture is trained for a single epoch (batch size 64, learning rate 0.01, momentum 0.9, dropout 0.2), and validation accuracy is recorded. The first approach is LLM-based: prompted with the desired interface and examples of Torchvision transforms, the model proposes augmentation functions as Python code, which are syntactically validated before integration. The second approach is combinatorial: from a predefined set of Torchvision transforms, all pipelines with one, two, or three variable transforms are enumerated and extended by a fixed suffix (\texttt{Resize}--\texttt{ToTensor}--\texttt{Normalization}); variable transforms receive randomly sampled parameters. This yields 6{,}000 unique augmentation pipelines, each evaluated under the same training protocol.

\begin{figure}
    \centering
    \includegraphics[width=0.5\linewidth]{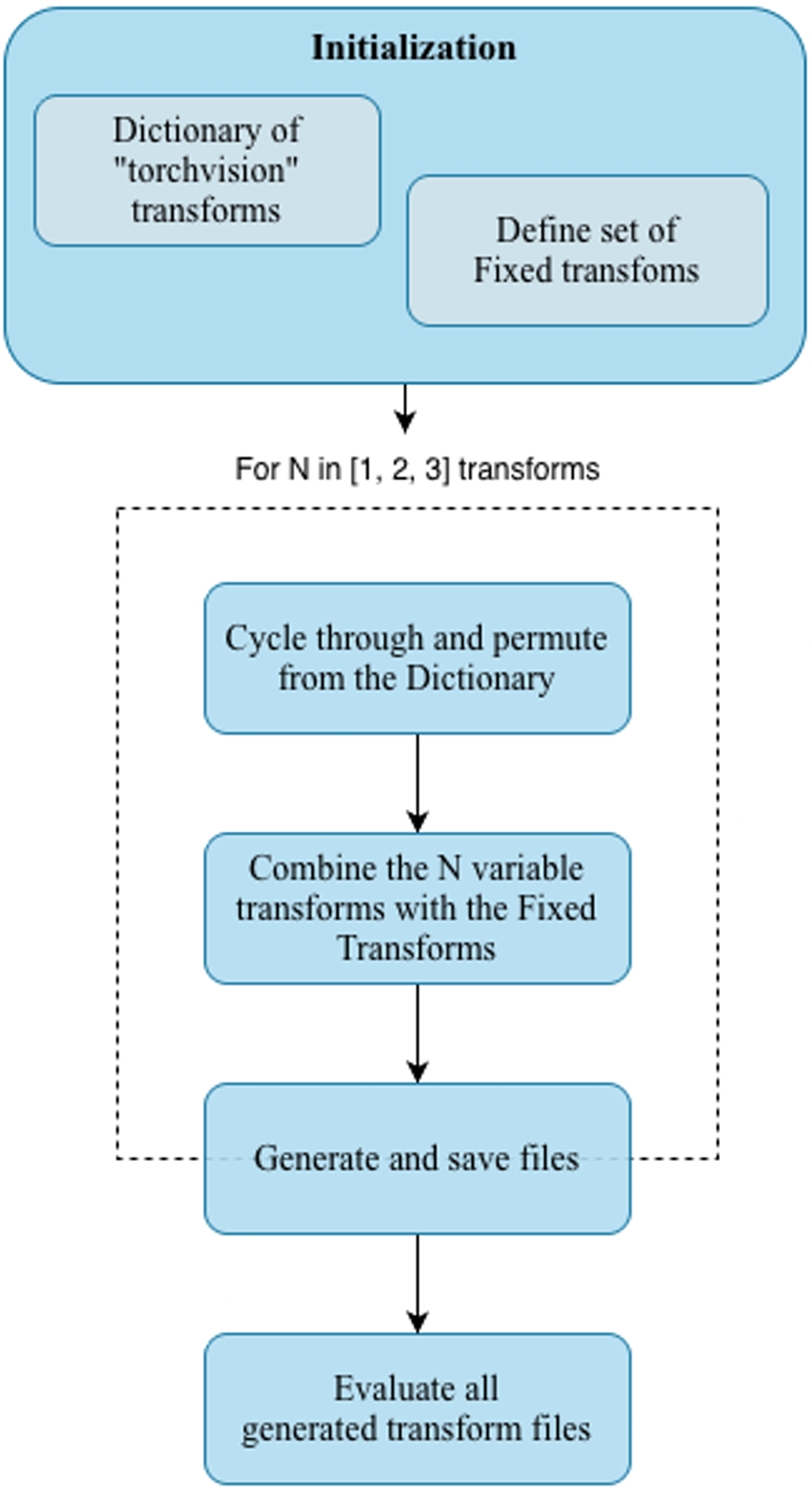}
    \caption{Overview of the brute-force generation of augmentation pipelines by permuting a fixed set of transforms and appending a standard preprocessing suffix.}
    \label{fig:transform_pipeline}
\end{figure}

\subsection{Cross Domain Deployment and Testing}

\subsubsection{NN-VR}

The VR-Ready Neural Network Verifier (NN-VR) provides automated validation and deployment of pre-trained LEMUR models in Unity-based virtual reality environments. Its architecture comprises (i) a Neural Network Parser that ingests ONNX exports and associated metadata from LEMUR and populates a Unity project configured with the Barracuda inference engine; (ii) a Compatibility Verifier that imports ONNX models, checks operator coverage and shader support, and profiles GPU memory under VR-relevant settings; and (iii) an Automated Porting System that converts and optionally optimizes models, configures inference scenes, and records diagnostic logs. Performance is evaluated against VR-specific criteria: inference latency relative to an 11\,ms-per-frame budget (90\,Hz), additional memory overhead compared to standalone inference, and numerical consistency between Unity/Barracuda outputs and reference PyTorch predictions within a tight tolerance. The pipeline is designed for scalable application to large collections of models exported from LEMUR.

\begin{figure}
    \centering
    \includegraphics[width=1\linewidth]{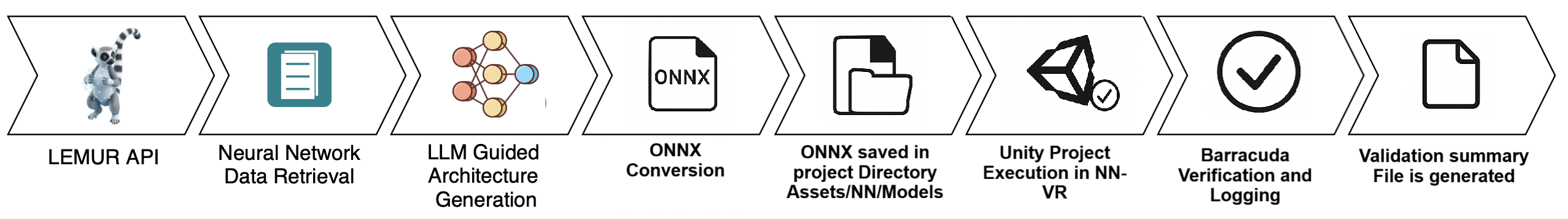}
    \caption{NN-VR pipeline}
    \label{fig:nnvr_pipeline}
\end{figure}

\subsubsection{NN-Lite}

NN-Lite~\cite{ABrain.NN-Lite} is an automated pipeline for deploying and benchmarking PyTorch models on the Android platform. It provides an end-to-end workflow designed to evaluate models from the dataset, managing the process from model conversion to final reporting without manual intervention.

The pipeline executes a four-stage process for each model. First, PyTorch models are automatically converted to the TensorFlow Lite (TFLite) format. This stage includes a custom wrapper to resolve the NCHW (PyTorch) to NHWC (TFLite) tensor layout disparity, ensuring compatibility and efficient memory usage. Second, the system manages the lifecycle of an Android Virtual Device (AVD), including boot-up and state monitoring. Third, a lightweight Android application is deployed to the AVD to execute model inference and collect latency metrics. Fourth, the pipeline retrieves these benchmark results and combines them with device analytics (e.g., memory, CPU architecture) into a structured JSON report.


The system is designed for large-scale, continuous operation, with built-in state management and failure recovery mechanisms. In a documented 48-hour session, the pipeline processed over 7,500 models from the LEMUR dataset. The final stage of the process is statistical consolidation, where performance metrics like task-level accuracy are aggregated with on-device latency data. This integration allows for a unified analysis of a model's software performance and its on-device execution characteristics.

\section{Evaluation}

All computer vision experiments are run in the AI Linux Docker image\footnote{AI Linux: \scriptsize \url{https://hub.docker.com/r/abrainone/ai-linux}} on NVIDIA GeForce RTX 3090/4090 GPUs with 24\,GB memory, using a Kubernetes cluster and dedicated workstations.

We performed repeated evaluations under multiple training configurations for every supported task and its associated datasets. Each evaluation run corresponds to a distinct hyperparameter setting, including batch size, number of training epochs, learning rate, momentum, dropout rate, and the selected data transformation pipeline. Every run is logged together with its accuracy and runtime, which enables systematic comparison across configurations and supports analysis of model performance variability under the same task protocol.

Evaluations use bounded training schedules to keep compute comparable while sweeping many architectures and hyperparameter settings. Metrics generally improve with additional optimization and data exposure, but convergence rates vary across configurations, so reported scores reflect compute-budgeted performance rather than uniformly converged endpoints. Accordingly, comparisons to fully trained baselines should use matched training horizons, and the results primarily quantify how efficiently each method converts compute into task performance.

\section{Results}
\begin{figure}
    \centering
    \includegraphics[width=1\linewidth]{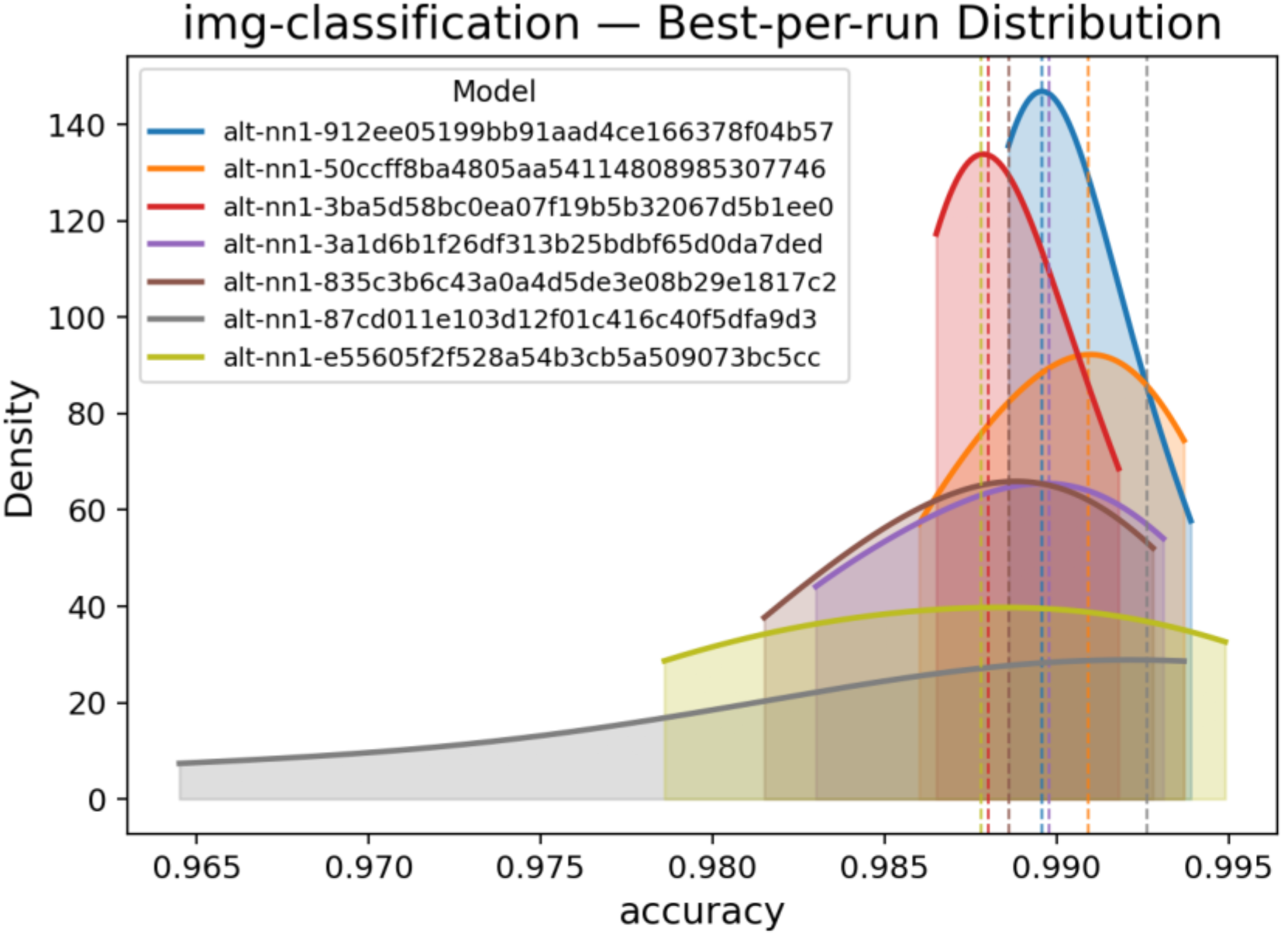}
    \caption{The distribution of the Few-Shot Architecture Prompting generated networks tested on MNIST.}
    \label{fig:best_per_run_classification}
    \vspace{-3.2mm}
\end{figure}

Table~\ref{tab:lemur2} summarizes the quantitative performance of the LEMUR~2 generative systems and deployment pipelines under bounded training budgets. Across the evaluated settings, optimization-guided generators produce the strongest and most consistent image-classification results, while prompt-based and programmatic mutation methods exhibit higher variance in both quality and reliability. Best-per-run denotes the highest accuracy attained across all training epochs for a fixed hyperparameter configuration, allowing comparison of peak observed performance rather than final-epoch values as seen in \ref{fig:best_per_run_classification}, \ref{fig:BLUE-4}.

\begin{figure}
    \centering
    \includegraphics[width=1\linewidth]{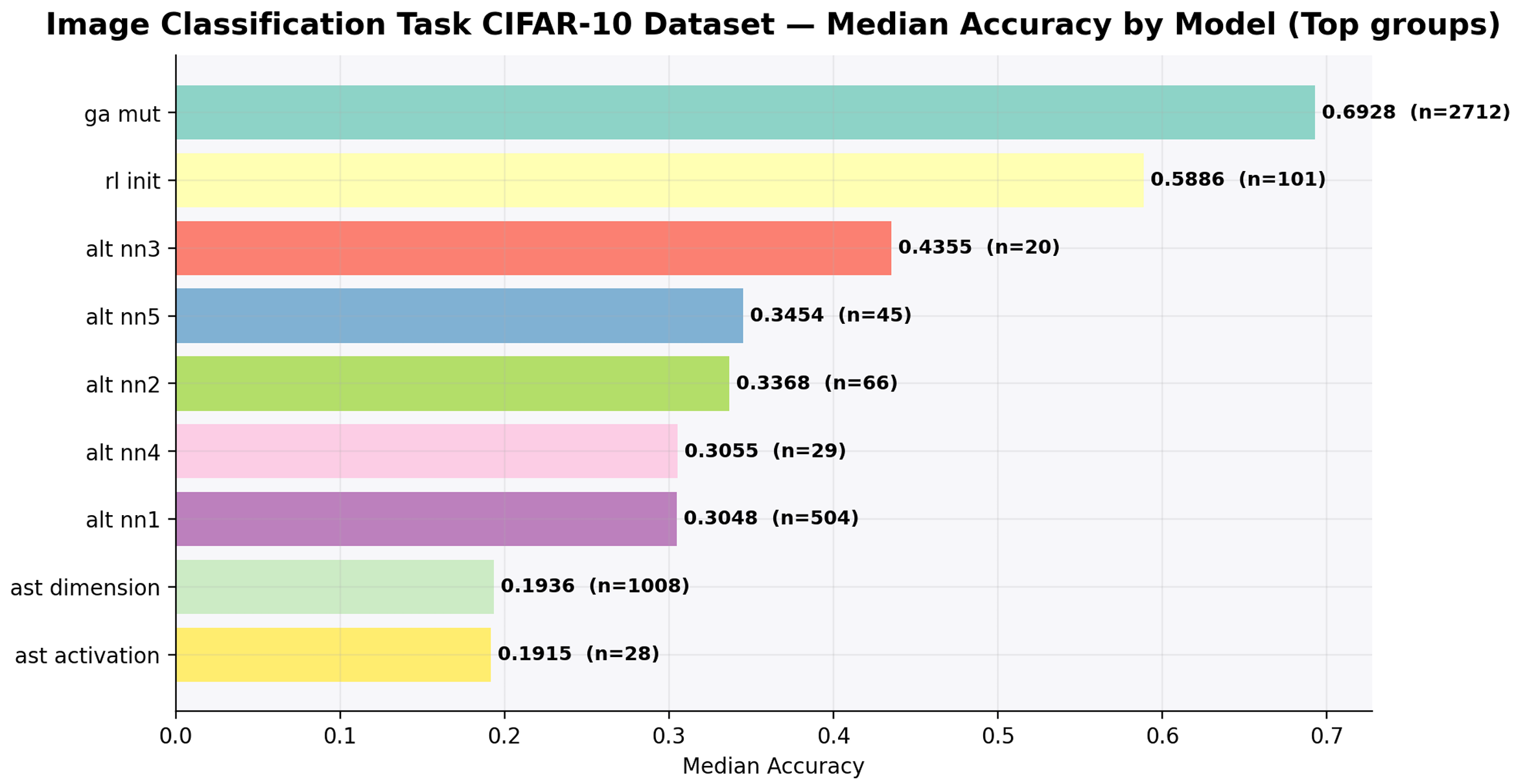}
    \caption{Median accuracy per model group for image classification task on the CIFAR-10 dataset.}
    \label{fig:median_cifar10}
\vspace{-3.2mm}
\end{figure}

Overall, these generative mechanisms produce models that are competitive with existing manually crafted network baselines. Within groups (Fig.~\ref{fig:median_cifar10}), evolutionary models attain the highest median accuracy, indicating a dense cluster of strong architectures. LLM-generated few-shot models (``alt-'') occasionally match the best groups at the top but have lower median accuracy and higher variance, while programmatic AST mutations (``ast-'') yield the lowest median accuracy, consistent with the sensitivity of local channel edits to the choice of input model relative to global optimization strategies.

This behavior can be explained by considering the effective size of the search space. Optimization-based generators converge faster than pure random search because the objective function provides a direction that quickly focuses the search on promising regions, especially when the coverage is moderate and the landscape is relatively smooth. As the problem becomes more complex and the search space grows, however, the landscape becomes increasingly rugged, making it harder for any method to consistently find high-quality solutions. The impact of architectural configuration, such as channel widths in the AST channel-mutated AlexNet variants show different convergence properties when compared with the original AlexNet, despite being a relatively small modification, underscoring how sensitive convergence can be to seemingly minor configurational change.

\begin{table*}[t]
\centering
\caption{Quantitative summary of LEMUR 2 generative systems and deployment pipelines.}
\label{tab:lemur2}
\scriptsize
\setlength{\tabcolsep}{3.5pt}
\begin{tabular}{@{}l@{\hspace{3pt}}l@{\hspace{4pt}}r@{\hspace{6pt}}l@{\hspace{6pt}}l@{\hspace{6pt}}r@{\hspace{6pt}}>{\centering\arraybackslash}p{2.6cm}@{\hspace{6pt}}>{\raggedright\arraybackslash}p{4.2cm}@{}}
\toprule
\textbf{Method} & \textbf{Prefix} & \textbf{\#} & \textbf{Dataset} & \textbf{Metric} & \textbf{Best Performance} & \textbf{Success Rate} & \textbf{Best / Notable Configuration} \\
\midrule
Genetic Algorithm              & ga-   & 2000 & CIFAR-10  & Accuracy & 0.8004        & 100\%   & Block-level evolution (pooling, act., BN) \\
Few-Shot LLM Prompting         & alt-  & 4033 & CIFAR-10  & Accuracy & 0.3874$^{a}$  & --- & 1 exemplar ($n$=1) best on CIFAR-10 \\
Fractal Networks               & frac- & 1258 & CIFAR-10  & Accuracy & 0.8018$^{b}$  & 97\%   & Recursive multi-column + AMP \\
AST Channel Mutation           & ast-  & 1129 & CIFAR-100 & Accuracy & 0.3110        & 100\%   & Non-standard widths + Late-stage exp. \\
Reinforcement Learning (GRPO)  & rl-   & 512  & CIFAR-10  & Accuracy & 0.6799$^{a}$  & 60\%    & Masked skeleton completion \\
NN-RAG                         & rag-  & 1289 & CIFAR-10  & Accuracy & 0.9281        & 73.0\%  & timm + torchvision + transformers \\
Data Augmentation (Brute Force)& ---   & 6000 & CIFAR-10  & Accuracy & 0.6124$^{a}$  & 100\%   & RandomPosterize + standard suffix \\
Data Augmentation (LLM-Gen)    & ---   & 280  & CIFAR-10  & Accuracy & 0.5728$^{a}$  & 22\%    & RRC + ColorJitter + Flip + Blur \\
\midrule
Mixture-of-Experts             & moe-  & 8    & CIFAR-10  & Accuracy & 0.9390        & 100\%   & Homogeneous top-2 routing (MoEv7) \\
                               &       &      &           &          & 0.9313        & 100\%   & Heterogeneous (Alex+Air+Dense+BagNet) \\
Image Captioning (NN-Caption)  & C*C   & 357  & MS-COCO   & BLEU-4   & 0.3170          & $>$50\% & ResNet-50 + Transformer decoder (768-dim) \\
Text-to-Image                  & t2i-  & 3    & ---       & CLIP Score & 0.2751       & 100\%   & CVAE-GAN + CLIP enc. + PatchGAN \\
\midrule
NN-Lite (Android TFLite)       & ---   & 7512 & ---       & Latency DB & ---          & 100\% conversion & NCHW$\rightarrow$NHWC wrapper \\
NN-VR (Unity/Barracuda 90 Hz)  & ---   & 10244& ---       & Frame Time & ---          & 95.0\% & Auto shader/memory optimization \\
\bottomrule
\end{tabular}

\vspace{2pt}
\parbox{\textwidth}{\footnotesize
$^{a}$ Tested on multiple datasets; the reported result corresponds to the dataset listed in the \textit{Dataset} column.
$^{b}$ Evaluated for 5 epochs.
}
\end{table*}

\subsection{Task Extensions}
\begin{figure} 
\centering 
\includegraphics[width=1.\linewidth]{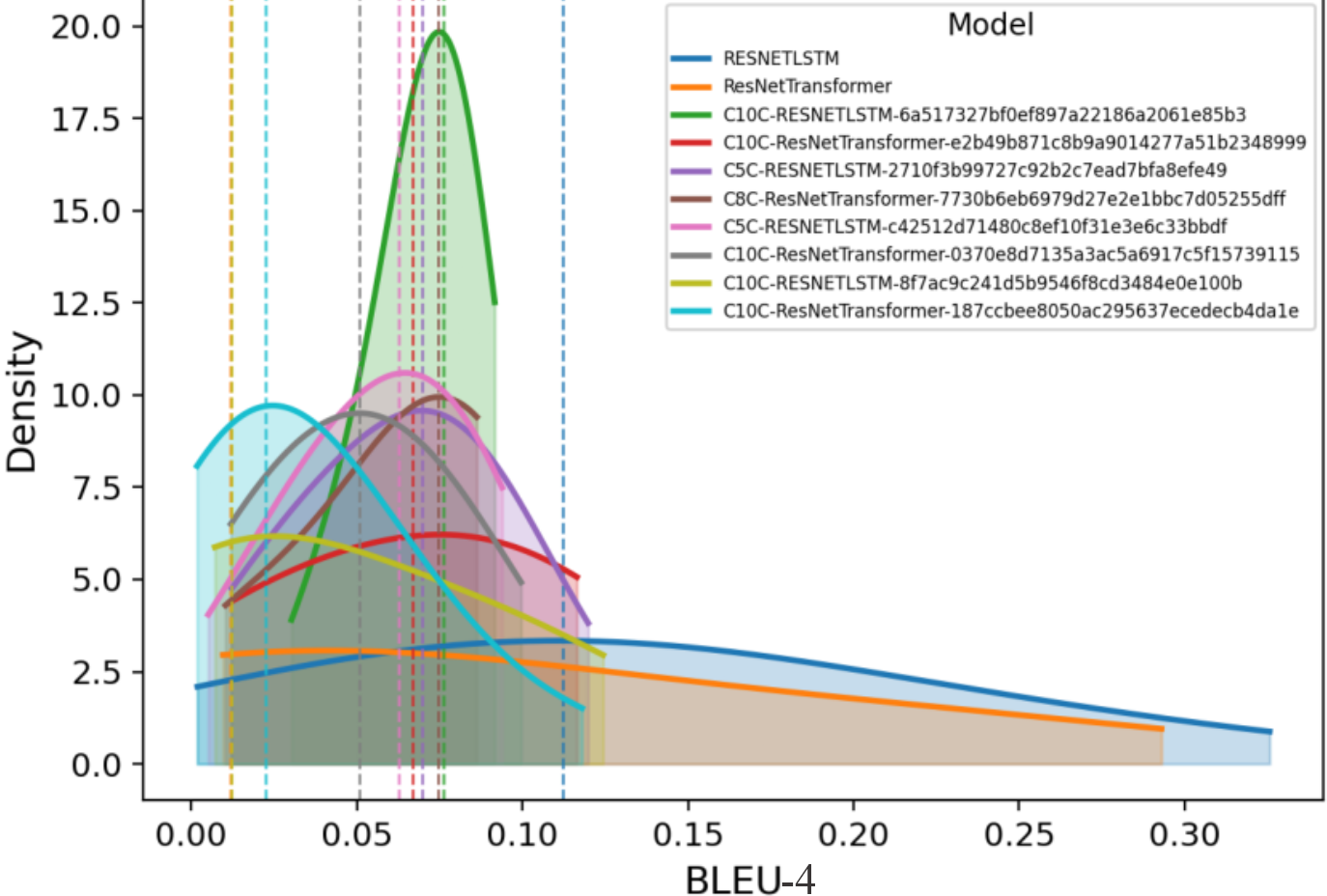} 
\caption{Best per-run image captioning BLEU-4 distributions.} 
\label{fig:BLUE-4} 
\vspace{-4mm} 
\end{figure}

\textbf{Image Captioning} Increasing prompt complexity (5--10 snippets) reduced runnable generations from 80\% to 50\%. The LLM explored diverse architectures (e.g., ConvNeXt, EfficientNet, Transformers); the best generated variant (ResNet-50 + Transformer) yielded a score of BLEU-4 = 0.317 at 50 epochs. In the longer training schedule, it can confidently compete with the baseline, ~(\cref{fig:BLUE-4}) approaching to the score of 0.3246. This highlights the trade-off between accuracy and architectural diversity.

\textbf{Text-to-Image} Methods progressed from a UNet diffusion model (CLIP 0.17--0.24) and LSTM-GAN ($\approx$0.214) to a complex CVAE-GAN. The final CVAE-GAN, utilizing a CLIP encoder, PatchGAN discriminator, and adversarial losses, achieved the peak CLIP score of 0.2751, prioritizing semantic alignment over photorealistic textures.

\textbf{Mixture-of-Experts} Evaluations on CIFAR-10 showed a tuned homogeneous MoE achieving the highest accuracy (93.9\%). A heterogeneous variant combining AlexNet, AirNet, DenseNet, and BagNet experts reached 93.13\%, successfully outperforming all constituent backbones trained individually.

\subsection{Generative Systems}
\textbf{Genetic Algorithm} A hyperparameter-restricted search yielded 62.76\% test accuracy (11{\texttimes}11 kernel), whereas evolving block-type structures (pooling, activation, BN) improved accuracy to 80.04\%. The champion model utilized batch normalization, 3{\texttimes}3 convolutions, and mixed pooling, demonstrating the value of block-level flexibility.

\textbf{NN-RAG} From 1,289 candidate blocks sourced from \texttt{timm}, \texttt{torchvision}, and \texttt{transformers}, 941 (73.0\%) passed execution checks. The resulting library, tagged ``rag-'' in LEMUR~2, includes executable attention, convolutional, and normalization modules.

\textbf{Few-Shot Prompting} Results were non-monotonic: $n=3$ exemplars achieved the highest balanced mean accuracy (53.1\%) considering all datasets and improved CIFAR-100 performance by 11.6pp ($p=0.001$). Conversely, $n=6$ caused a 99.8\% failure rate due to context overflow.

\textbf{Data Transformation} The best LLM-generated pipeline achieved 57.28\% validation accuracy. A combinatorial search of 6,000 pipelines identified a superior configuration using a single \texttt{RandomPosterize} transform, reaching 61.24\% accuracy and outperforming the generative approach.


\begin{figure}
    \centering
    \includegraphics[width=1\linewidth]{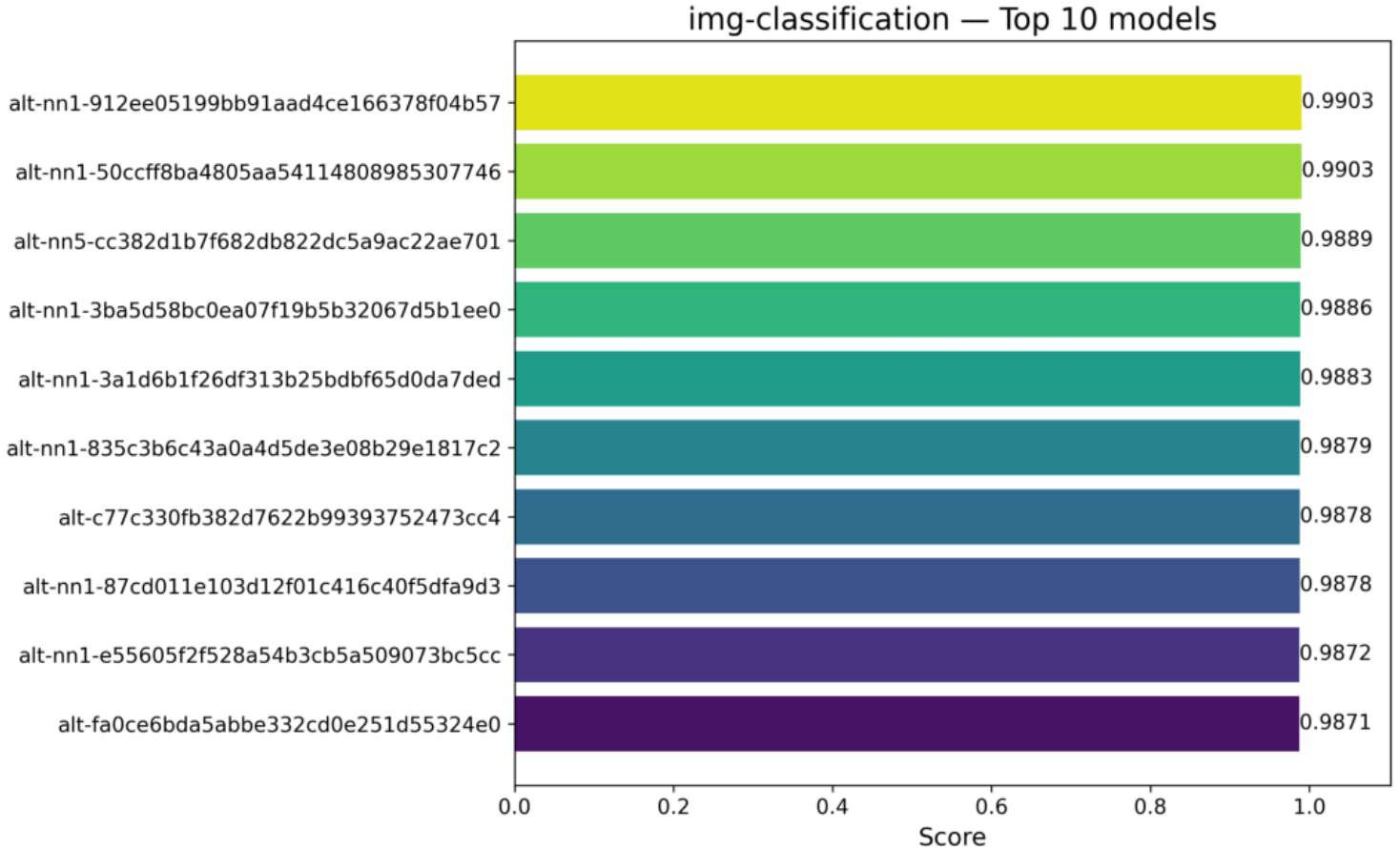}
    \caption{Top 10 image classification models across the dataset.}
    \label{fig:top_models_classification}
\end{figure}

\section{Conclusion}




This system was developed to produce a large and architecturally diverse collection of neural networks with extensive performance records across multiple tasks and hardware platforms, resulting in over 14{,}000 distinct architectures and more than 750{,}000 structured training records documenting model architectures, training configurations, and task-specific metrics such as accuracy. For edge devices and cross-domain coverage, NN-Lite and NN-VR automate model deployment and benchmarking while recording on-device characteristics including inference latency and memory usage. By mobilizing data at scale, the framework reveals patterns that indicate which network configurations and structures are effective, supporting automated development of next-generation architectures~\cite{ABrain.NNGPT}. By formalizing dependency-closed neural primitives, it also establishes a scalable methodology for ``Neural Architecture Mining,'' transforming disparate source code into a standardized, executable substrate for automated ML workflows. Due to space constraints, further implementation details, ablation studies, and extended results are deferred to the supplementary material.

\vspace{0.2cm}
\noindent\textbf{Acknowledgments.}
This work was partially supported by the Alexander von Humboldt Foundation.


{
    \small
    \bibliographystyle{ieeenat_fullname}
    \bibliography{main}
}
\end{document}